\documentclass[10pt,twocolumn,letterpaper]{article}

\usepackage{cvpr}              
\usepackage{booktabs}
\usepackage{algorithm}
\usepackage[noend]{algorithmic}
\usepackage[accsupp]{axessibility} 
\definecolor{cvprblue}{rgb}{0.21,0.49,0.74}
\usepackage[pagebackref,breaklinks,colorlinks,allcolors=cvprblue]{hyperref}
\usepackage{comment}
\def\paperID{33584} 
\def\confName{CVPR}
\def\confYear{2026}

\title{Language-Augmented Semantic Priors for B-Spline Surface Fitting}

\author{Yunzhong Lou \quad
Yusheng Luo \quad
Jiahao Li \quad
Yu Song \quad
Xiangdong Zhou* \\
College of Computer Science and Artificial Intelligence, Fudan University, Shanghai, China\\
{\tt\small 
\{yzlou20, xdzhou\}@fudan.edu.cn \quad
\{lijh23, ycluo24, songy23\}@m.fudan.edu.cn}
}

\begin{document}
\twocolumn[{
\renewcommand\twocolumn[1][]{#1}
\maketitle
\begin{center}
        \fontsize{10.25pt}{\baselineskip}\selectfont
\end{center}
\vspace{.4mm}
    \begin{center}
        \vspace{-10mm}
        \centering
        \includegraphics[width=1\linewidth]{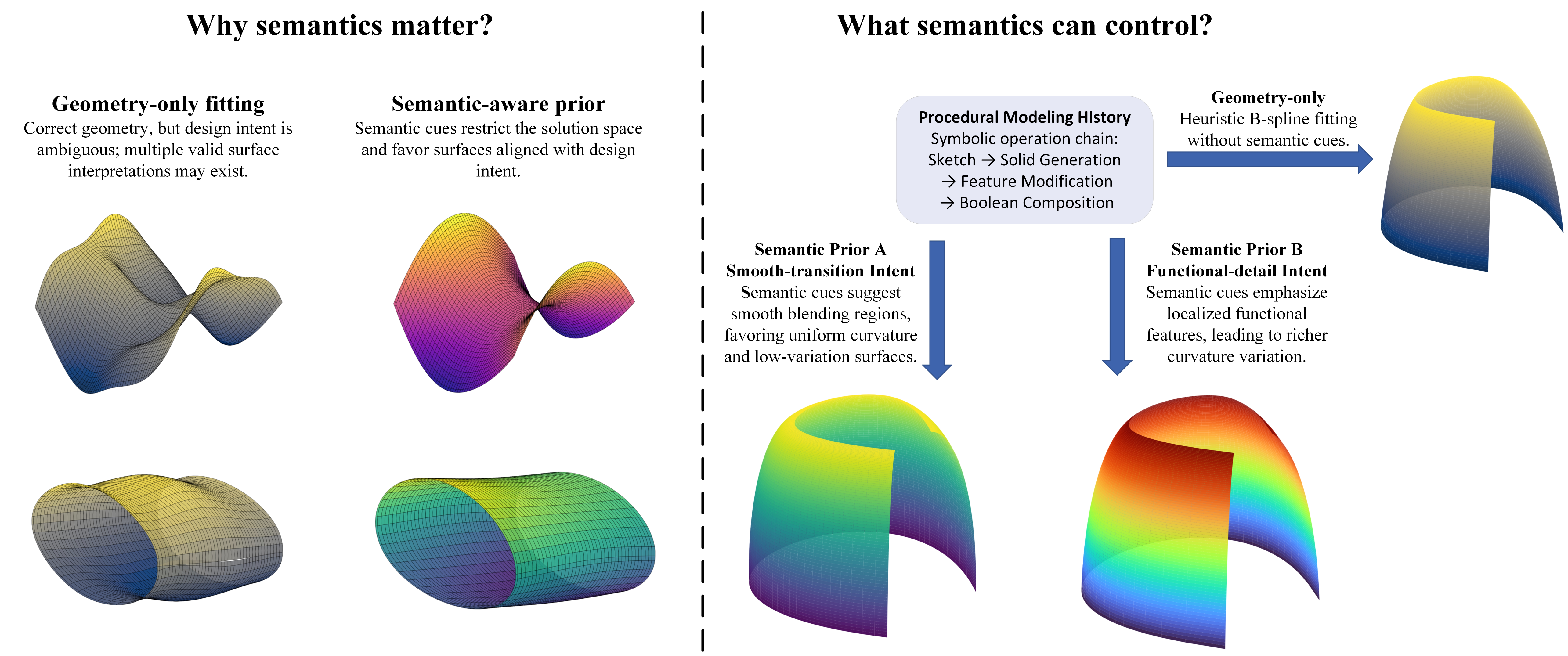}
        \vspace{-3mm}
        \captionof{figure}{\textbf{Semantic information meaningfully constrains the geometric interpretation of B-spline surfaces.} 
When relying solely on geometric cues, a solver may produce multiple equally valid but intent-agnostic surface configurations. 
Incorporating semantic cues—such as smooth-transition intent or functional-detail emphasis—reduces this ambiguity and steers the fitted surface toward characteristics aligned with design intent.
}
        \label{fig:figure1}
    \end{center}
}]
\begin{abstract}
The use of B-splines and Non-Uniform Rational B-Splines (NURBS) surfaces constitutes the mathematical foundation of contemporary three-dimensional computer-aided design (CAD) systems. Despite long-term progress, geometric kernels in traditional CAD still rely heavily on predetermined heuristic initialization for surface fitting and parameterization. Meanwhile, the rich procedural semantics and design intent encoded in modeling histories (i.e., sequences of design operations) are largely ignored during geometry generation. This disconnect creates a critical gap between high-level design intent and solver-executable geometric configuration, often leading to suboptimal and semantically inconsistent fitting results.  To bridge this gap, we introduce \textbf{LASP}, a \textit{Language-Augmented Semantic Priors} framework that leverages large language models (LLMs) to infer structured, solver-usable B-spline priors from procedural modeling histories. Rather than modifying the geometric kernel itself, LASP operates as a semantic reasoning layer above existing solvers. It first translates modeling histories into rich textual descriptions that capture design intent, geometric context, and functional relationships, and then uses a fine-tuned LLM to predict structured B-spline prior parameters. LASP is trained through a two-stage scheme that combines local geometric regularities with long-range contextual dependencies, producing priors that are both interpretable and semantically coherent. This approach furnishes enhanced inductive signals that direct the conventional B-spline fitting process toward solutions that more accurately encapsulate the intended design objectives and demonstrate heightened semantic coherence. Compared to traditional machine learning schemes, the experiments demonstrate that language-driven reasoning can serve as a powerful inductive bias for geometric solving, establishing a new paradigm of language-guided geometric optimization in modern CAD systems.
 
\end{abstract}
    
\section{Introduction}

B-spline and Non-Uniform Rational B-Splines (NURBS) surfaces provide the mathematical backbone of modern CAD systems~\cite{camba2016parametric,hoffmann1991fundamental,piegl2012nurbs}, supporting smooth, editable, and topologically consistent geometric representations. Commercial kernels such as Parasolid and ACIS, as well as open-source systems like OpenCascade~\cite{riegel2016freecad}, implement decades of carefully engineered numerical routines to ensure robustness and validity in geometric computation. Despite their maturity, these solvers depend on fixed heuristic initialization—predefined degrees, uniform knot vectors, and static control-point layouts—before executing deterministic fitting procedures. While effective for stability, such initialization limits the solver to a narrow region of its feasible parameter space and prevents it from adapting to the semantics of the modeling operation. Procedural commands such as \textit{Fillet}, \textit{Shell}, or \textit{Loft} encode rich information about curvature transitions, functional intent, and local geometry~\cite{willis2021fusion,wu2021deepcad,horvath2004treatise}, yet this knowledge is rarely translated into solver-usable geometric configurations for downstream surface generation.

Closing this gap requires connecting two fundamentally different representations. Modeling histories are symbolic, rule-based, and hierarchical, as captured in recent analyzes of parametric workflows~\cite{koch2019abc}; geometric solvers, by contrast, operate over continuous numeric manifolds. This structural mismatch makes it difficult for solvers to exploit procedural context, even though such context often determines the intended geometric behavior. Prior learning–based approaches typically regress geometric parameters or reconstruct surfaces directly~\cite{wang2025text,yin2025shapegpt,yang2023large}, leaving the solver unchanged and isolated from semantic guidance. Existing AI4CAD studies have predominantly focused on modeling-history reconstruction, or generating modeling histories and geometric outputs from design intent~\cite{lou2023brep,ma2024draw}, while the problem of translating designer-side semantics, or practical semantic proxies of them, into geometry-governing and solver-usable configurations remains largely underexplored. A more principled direction is therefore to endow solvers with context-conditioned priors that guide their initialization and restrict their search space toward semantically meaningful solutions—while retaining their deterministic optimization pipelines.

To address this challenge, we introduce \textbf{LASP} (\textit{Language-Augmented Semantic Priors}), a framework that leverages large language models (LLMs)~\cite{brown2020language,chowdhery2023palm} to infer semantically coherent B-spline priors from procedural modeling histories. LASP translates symbolic operations into rich-text semantic descriptions that capture multi-level cues about operation intent, geometric context, and functional relationships, aligning with recent language-driven CAD reasoning systems~\cite{xu2024cad,wang2025cad,xiaorui2025llms}. A fine-tuned LLM then predicts structured surface priors—degrees, control-point grids, rationality, closure, trimming patterns, continuity class, and surface roles—forming a language-conditioned parameterization of B-spline surfaces. Importantly, LASP does not modify the solver itself or introduce a new geometric optimization method; instead, it operates as a semantic reasoning layer above existing geometric kernels, configuring the fitting process through solver-usable priors and biasing the numerical optimization toward geometrically valid and semantically consistent regions of the parameter space.

LASP is trained in two complementary stages. Stage~I focuses on local geometric regularities by learning operation-specific parameter structures grounded in physically valid B-spline distributions~\cite{koch2019abc}. Stage II extends this reasoning to full modeling histories, capturing long-range dependencies such as continuity propagation, curvature transitions, and semantic role consistency across operations. This aligns with recent advances showing that LLMs can model
long-range symbolic dependencies~\cite{chen2021evaluating,zitkovich2023rt,poole2022dreamfusion}. Together, these stages enable the model to learn both geometric grounding and high-level design semantics.

Extensive experiments demonstrate that LASP substantially improves parameter prediction accuracy and semantic consistency over strong LLM baselines and Transformer-only architectures. We further conduct an engine-level study by mapping discrete semantic priors into solver-accessible configuration choices, showing that semantically guided parameterization significantly improves RMS, Hausdorff, and median fitting errors—highlighting the practical influence of symbolic semantics on geometric fitting behavior and complementing recent language-conditioned optimization pipelines~\cite{yao2023tree,schick2023toolformer}. Our key contributions are summarized as follows:
\begin{itemize}
    \item We show that semantic priors learned from procedural modeling histories serve as transferable inductive biases for geometric solvers, guiding numerical optimization toward semantically meaningful solutions without modifying solver internals.    
    \item We present \textbf{LASP}, the first language-driven framework that predicts structured B-spline priors from symbolic modeling histories via rich-text semantic reasoning and a two-stage LLM-based learning pipeline.
    \item We demonstrate that semantic priors significantly improve both parameter-level prediction and solver-level reconstruction quality, providing quantitative and qualitative evidence that linguistic reasoning provides a strong inductive bias for intent-aware geometric configuration.
\end{itemize}

\section{Related Work}
\begin{figure*}[t]
\vspace{-3mm}
    \centering
    \includegraphics[width=\linewidth]{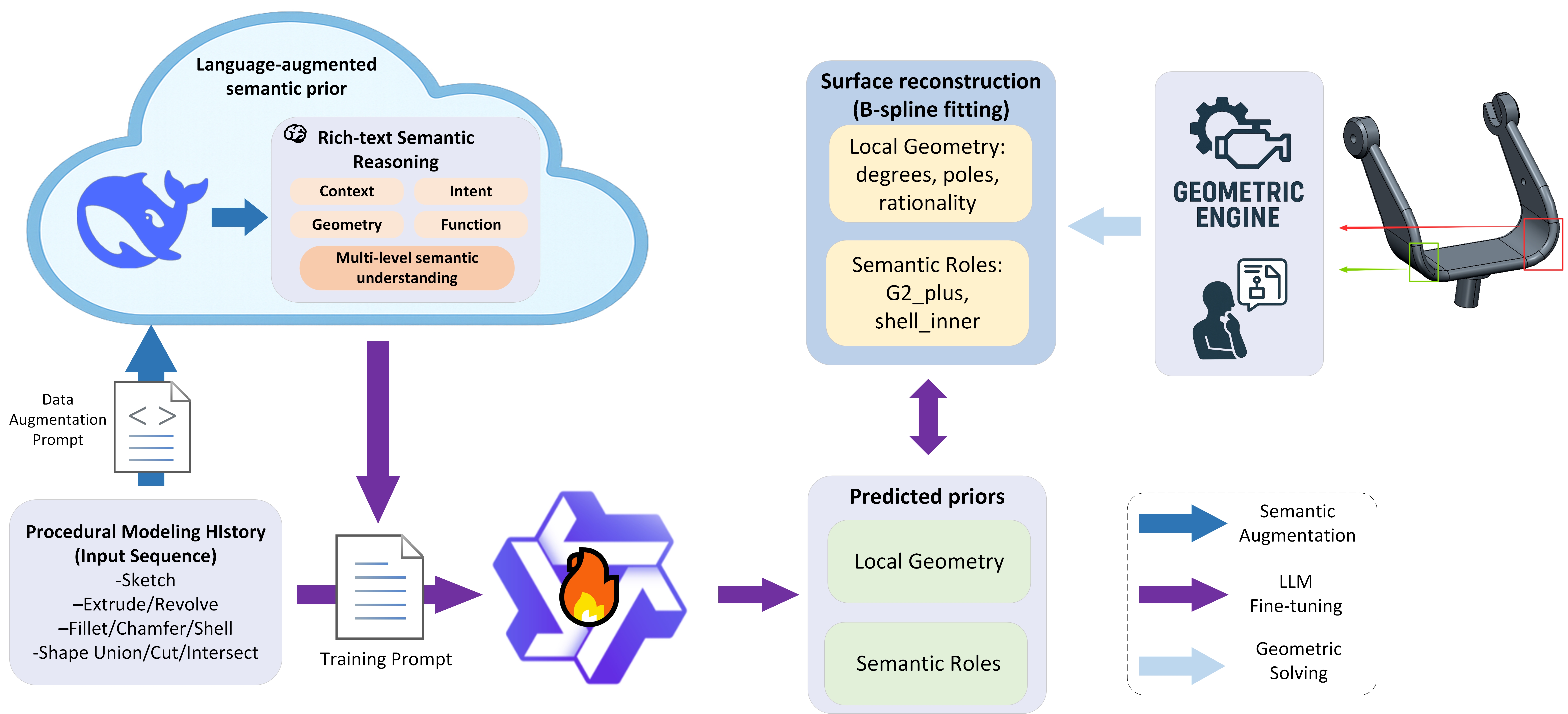}
    \caption{
    \textbf{LASP framework.}
    During training, given a procedural modeling history $H$, LASP first performs rich-text
    semantic reasoning to generate contextual descriptions $R$ that capture
    operation intent, geometric characteristics, and functional cues. The
    combined input $(H,R)$ is formatted into prompts and used to fine-tune a
    large language model to predict structured B\text{-}spline semantic priors
    $\mathcal{P}$, including local geometry and
    high-level surface roles. These priors serve as semantic initialization for
    downstream B\text{-}spline fitting within a geometric engine, guiding the
    solver toward semantically consistent surface reconstruction.}
    \label{fig:framework}
    \vspace{-5mm}
\end{figure*}

\subsection{LLMs for Structured and Symbolic Reasoning}
Large language models have shown strong capabilities in processing structured and symbolic data beyond natural language~\cite{brown2020language,chowdhery2023palm}. Transformer-based models have been successfully applied to program synthesis, code completion, and symbolic querying~\cite{qin2022survey}, with systems such as Codex~\cite{chen2021evaluating}, CodeT5~\cite{wang2021codet5}, and SQL-PaLM~\cite{sun2023sql} demonstrating that pretrained LLMs can capture discrete dependencies and multi-step reasoning patterns directly from serialized sequences~\cite{wei2022chain}. Tasks involving text-to-SQL translation and knowledge-graph question answering further highlight that LLMs generalize well to symbolic domains when the underlying structure is appropriately encoded~\cite{yu2018spider,jin2024large}.

Motivated by these findings, we treat procedural modeling histories as structured symbolic sequences. When enriched with contextual descriptions, they allow LLMs to infer geometric intent and operational semantics that conventional numerical solvers cannot access.

\subsection{Semantic Modeling and Data-Driven CAD}
Semantic enrichment of CAD representations has long been a challenge~\cite{stroud2006boundary}. Classical formats such as BRep and CSG provide explicit geometry and topology but lack high-level semantic interpretability~\cite{bidarra2000semantic,brunetti2000feature}. Efforts such as ShapeNet~\cite{chang2015shapenet} semantics and ontology-driven design models~\cite{furini2016knowledge} aim to attach human-readable meaning to geometric entities, though typically through manual or rule-based processes with limited scalability.

Recent data-driven approaches have shifted CAD research toward learning from parametric workflows. DeepCAD~\cite{wu2021deepcad} models distributions over CAD command sequences, while CAD-Llama~\cite{li2025cad} extends this direction with large language models for parametric CAD generation. Seek-CAD~\cite{li2025seek} and ReCAD~\cite{li2026recad} further explore self-refinement, multimodal reasoning, and reinforcement-enhanced generation in text or image conditioned CAD modeling. However, these methods primarily aim at generating CAD programs or modeling sequences, rather than inferring interpretable semantic priors that can directly configure downstream geometric solvers. 

\vspace{-3pt}
\subsection{B-Spline Parameterization and Geometric Solving}
B-spline and NURBS surfaces~\cite{piegl2012nurbs} remain the foundation of modern CAD kernels due to their mathematical flexibility and robustness~\cite{hasan2024b,yang2023topology}. Solvers such as OpenCascade~\cite{banovic2018algorithmic}, Parasolid~\cite{kramer1991extracting}, and ACIS typically rely on fixed initialization schemes (uniform knots, predefined degrees) followed by deterministic least-squares optimization~\cite{yuan2021computing}. Although effective for geometric validity, these procedures operate independently of their procedural or functional context.

Prior efforts to improve B-spline solving—such as curvature-aware refinement, adaptive pole allocation, or energy-based fitting~\cite{zhang2016b}—remain heuristic and local. In contrast, we inject semantic priors inferred by an LLM to guide solver initialization, enabling more meaningful convergence behavior without modifying kernel internals.

\subsection{Hybrid Language--Geometry Reasoning}
Hybrid reasoning frameworks that integrate language models with continuous numerical systems have shown promising results across generalist agents~\cite{reed2022generalist}, molecular modeling~\cite{kang2024chatmof}, constraint solving~\cite{shi2025constraintllm}, and text-driven 3D synthesis~\cite{zitkovich2023rt}. These works demonstrate that textual cues can shape optimization trajectories and solution spaces.

Inspired by this paradigm, recent CAD-oriented studies~\cite{xiaorui2025llms} explore the connection between symbolic reasoning and geometric processing. Our framework advances this direction by introducing structured semantic priors that bridge procedural semantics and B-spline parameterization, enabling solver-compatible guidance where language provides global intent and the geometric kernel enforces local validity.

\section{Method}
\subsection{Overview and Problem Definition}
\label{subsec:overview}
We propose \textbf{LASP} (\textit{Language-Augmented Semantic Priors}), a language-driven framework that bridges procedural modeling and geometric reasoning through contextual supervision.  
An overview of the framework is illustrated in Fig.~\ref{fig:framework}.  
LASP introduces a semantic reasoning layer between symbolic modeling and numerical solving, 
learning to infer \textit{semantic priors} that parameterize B-spline surface generation.  
This design enables contextual understanding of geometric operations 
while remaining fully independent of solver internals.

\vspace{-5pt}
\paragraph{B-spline Surface Fitting.}
Given control points $\{P_{ij}\}$, knot vectors, and basis functions $N_{i,p}(u)$ and $N_{j,q}(v)$, 
a tensor-product B-spline surface is defined as\cite{piegl2012nurbs}:
\begin{equation}
\label{eq:bspline_def}
S(u,v) = \sum_{i=0}^{N_u}\sum_{j=0}^{N_v} N_{i,p}(u)\, N_{j,q}(v)\, P_{ij},
\end{equation}
where $(p,q)$ denotes the polynomial degrees and $(N_u,N_v)$ represents the control-point grid.  
Conventional geometric kernels determine these parameters heuristically and 
fit the surface by minimizing the least-squares energy\cite{ma1995parameterization}:
\begin{equation}
\label{eq:bspline_fit}
E_{\text{fit}} = \sum_{k}\|\,\mathbf{x}_k - S(u_k,v_k)\,\|^2,
\end{equation}
which is highly sensitive to initialization and parameterization.  
LASP aims to infer a structured set of surface parameters — including polynomial degrees $(p,q)$, 
control-point grids $(N_u,N_v)$, rational flags $(\rho_u,\rho_v)$, and semantic attributes such as trimming and continuity — 
which act as semantic priors regularizing the fitting process and embedding design intent 
directly into the parameter space of geometric optimization.

\vspace{-5pt}
\paragraph{Symbolic and Linguistic Representations.}
A procedural modeling history is represented as an ordered sequence of parametric operations:
\begin{equation}
\label{eq:history_def}
H=\{o_t=(\mathrm{type}_t,\mathrm{param}_t)\}_{t=1}^{T},
\end{equation}
where $\mathrm{type}_t$ denotes the operation category (\textit{Fillet}, \textit{Shell}, \textit{Chamfer}, etc.), 
and $\mathrm{param}_t$ specifies its geometric arguments.  
To enrich these symbolic operations with contextual, functional, and geometric semantics,  
a \textit{rich text semantic representation} 
\[
R=\{R_t\}_{t=1}^{T}
\]
is generated by a language reasoning module (corresponding to the ``Rich-text Semantic Reasoning'' block in Fig.~\ref{fig:framework}).  
Each $R_t$ captures multi-level cues about design context, intent, geometry, and function,  
serving as the linguistic complement to the symbolic history $H$.

\vspace{-5pt}
\paragraph{Semantic Prior Prediction Task.}
Given a symbolic history $H$ and its corresponding rich-text representation $R$,  
LASP learns to predict a structured prior for each operation:
\begin{equation}
\label{eq:prior_def}
\begin{split}
P_t = \{&d_u,d_v,N_u,N_v,\rho_u,\rho_v,\\
        &\mathrm{trim},\mathrm{closed}_u,\mathrm{closed}_v,
        \mathrm{cont},\mathrm{curv},\mathrm{role}\}.
\end{split}
\end{equation}
Each $P_t$ encodes both numerical and semantic attributes—such as polynomial degrees, control-point counts, 
rationality, trimming, continuity, and surface roles.  
The complete prior sequence $\mathcal{P}=\{P_t\}_{t=1}^{T}$ 
encapsulates geometric regularities and high-level semantic relations throughout the modeling process.  
Detailed field definitions and examples are provided in the supplementary material.

\vspace{-5pt}
\paragraph{Generative Mapping and Training Objective.}
LASP learns a generative mapping:
\begin{equation}
\label{eq:llm_mapping}
f_\theta : (H, R) \rightarrow \mathcal{P}=\{P_t\}_{t=1}^{T},
\end{equation}
where $f_\theta$ is parameterized by $\theta$ and implemented using an instruction-tuned large language model.  
During fine-tuning, the model receives concatenated symbolic histories and their rich-text semantic descriptions as input and autoregressively generates structured priors in a text-to-structured-text format.  
Training is carried out in two stages (Sec.~\ref{subsec:prior}), 
but both stages share the same token-level objective, namely the negative log-likelihood (NLL):
\begin{equation}
\label{eq:llm_nll}
\mathcal{L}_{\text{NLL}}
= - \sum_{t=1}^{T}\sum_{i \in \mathcal{I}_t}
\log p_{\theta}\!\left(y_{t,i}\mid y_{t,<i},\,H, R\right),
\end{equation}
where $\mathcal{I}_t$ indexes valid output tokens for the $t$-th operation.  
This unified objective implicitly enforces both linguistic and geometric consistency, 
as each target sequence jointly encodes semantic reasoning and B-spline parameter values.

In summary, LASP formulates B-spline parameter inference as a language-conditioned generative process, 
where each procedural operation is enriched with multi-level semantic reasoning $R$ and optimized through the autoregressive objective in Eq.~\ref{eq:llm_nll}.

\subsection{Semantic Prior Prediction}
\label{subsec:prior}

Building on the generative mapping in Sec.~\ref{subsec:overview}, 
LASP learns to infer semantic priors $\mathcal{P}=\{P_t\}_{t=1}^{T}$ that encapsulate 
the structural configuration of B-spline surfaces generated during procedural modeling.  
Each $P_t$ describes the expected parameterization pattern of the operation $o_t$ 
(e.g., degrees, control-point counts, continuity, trimming, and surface role), 
serving as a high-level inductive bias for downstream geometric solving.  
To capture both local parameter regularities and long-range semantic dependencies across the sequence, 
we adopt a two-stage conditional generative modeling scheme:
\[
p_{\theta}(\mathcal{P}\,|\,H,R)
=\prod_{t=1}^{T}p_{\theta}(P_t\,|\,P_{<t},H,R),
\]
where $H$ denotes the symbolic modeling history (design history), and $R$ its rich-text semantic enrichment.
A detailed prompt template, together with representative rich-text examples for $R$
, is provided in the supplementary material, illustrating how symbolic operations
are expanded into structured semantic descriptions.

\begin{table*}[t]
\centering
\caption{\textbf{Parameter-level prediction results.}
Comparison across geometric, topological, and semantic B-spline attributes.
Regression metrics use MAE; classification uses Accuracy and Macro-F1. 
Baselines include Transformer~\cite{vaswani2017attention}, GPT-5~\cite{achiam2023gpt}, and DeepSeek-v3.2~\cite{liu2024deepseek}.}
\label{tab:param_results}
\begin{tabular}{lcccc}
\toprule
\textbf{Metric} & \textbf{Transformer} & \textbf{GPT-5} & \textbf{DeepSeek-v3.2} & \textbf{LASP (Ours)} \\
\midrule
\multicolumn{5}{l}{\textbf{Geometric Parameters}} \\
\quad BSpl-Det (F1↑) & 0.507 & 0.559 & 0.667 & \textbf{0.950} \\
\quad Deg-Acc (U/V/Mean↑) & 0.436 / 0.416 / 0.429 & 0.520 / 0.501 / 0.507 & 0.622 / 0.593 / 0.608 & \textbf{0.883 / 0.824 / 0.854} \\
\quad Pole-MAE (U/V/Total↓) & 0.474 / 0.780 / 0.624 & 0.364 / 0.630 / 0.494 & 0.367 / 0.660 / 0.517 & \textbf{0.353 / 0.612 / 0.485} \\
\quad Rat-F1 (U/V↑) & 0.143 / 0.410 & 0.202 / 0.481 & 0.270 / 0.645 & \textbf{0.446 / 0.941} \\
\midrule
\multicolumn{5}{l}{\textbf{Topology}} \\
\quad Topo (Acc / F1↑) & 0.593 / 0.070 & 0.606 / 0.096 & 0.710 / 0.141 & \textbf{0.967 / 0.233} \\
\midrule
\multicolumn{5}{l}{\textbf{Semantic Attributes}} \\
\quad Cont (MF1 / Acc↑) & 0.280 / 0.430 & 0.380 / 0.489 & 0.478 / 0.596 & \textbf{0.698 / 0.824} \\
\quad Curv (MF1 / Acc↑) & 0.121 / 0.385 & 0.174 / 0.460 & 0.222 / 0.554 & \textbf{0.327 / 0.765} \\
\quad Role (MF1 / Acc↑) & 0.528 / 0.588 & 0.607 / 0.625 & 0.716 / 0.730 & \textbf{0.979 / 0.985} \\
\bottomrule
\end{tabular}
\vspace{-4mm}
\end{table*}

\vspace{-5pt}
\paragraph{Stage I: Local Prior Learning.}
The first stage focuses on capturing local geometric regularities that govern B-spline formation,
independent of semantic intent.
Given operations that directly produce new surfaces (e.g., \textit{Fillet}, \textit{Chamfer}, \textit{Shell}),
the model estimates a geometry-only prior distribution
\[
p_{\theta_1}(P_t\,|\,o_t),
\]
where $o_t$ denotes an isolated procedural command.
Each training pair $(o_t, P_t^{GT})$ supervises numerical attributes
$(d_u,d_v,N_u,N_v,\rho_u,\rho_v)$,
with small controlled perturbations for regularization.
This stage enables the model to internalize transferable structural laws of B-spline generation,
providing a stable inductive bias for subsequent semantic modulation.

\vspace{-5pt}
\paragraph{Stage II: Global Contextual Prior Learning.}
Building upon Stage~I representations,  
the second stage extends prior estimation to full modeling sequences,  
where operations interact through shared topology and design intent.  
The model learns
\[
p_{\theta_2}(P_t\,|\,P_{<t},H,R),
\]
introducing autoregressive dependencies among surface priors.  
Initialization with $\theta_1$ preserves low-level geometric consistency  
while the model learns higher-order relations—%
such as continuity propagation, curvature transitions, and semantic role coherence (e.g., ``transition fillet'' or ``shell offset'').  
This stage enables LASP to reason over long-range structural dependencies  
and infer consistent prior distributions across the procedural hierarchy.

\vspace{-5pt}
\paragraph{Prompt-Conditioned Training.}
Both stages are trained under the unified autoregressive objective in Eq.~\ref{eq:llm_nll},  
with stage-specific prompt schemas that emphasize different reasoning scopes.  
Stage~I prompts highlight local parameter inference and structural precision,  
whereas Stage~II prompts incorporate richer contextual cues  
to express inter-operation relationships and semantic continuity.  
Although the underlying optimization objective remains identical,  
schema-specific masking and conditioning provide field-aware supervision,  
aligning geometric precision with semantic coherence  
within a unified generative representation.

\vspace{-5pt}
\paragraph{LLM-based Implementation.}
The mapping $f_{\theta}$ is realized by fine-tuning an instruction-tuned large language model using a text-to-structured-text formulation.  
The model receives procedural histories augmented with their rich-text representations $R$ as input  
and outputs structured JSON-like prior sequences.  
Stage-wise prompt conditioning gradually shifts the model’s focus  
from local numeric regularity to broader procedural-semantic context,  
allowing LASP to generate interpretable and context-aware B-spline priors  
that are more consistent with geometric constraints and high-level semantic context.

\subsection{Theoretical Influence of Semantic Priors on Geometric Solving}
\label{subsec:solver}
The semantic priors predicted by LASP are not directly optimized inside the geometric kernel.  Instead, they act as structured biases over solver-accessible configuration choices,  such as polynomial degree, closure status, and trimming-related patterns.  From a conceptual probabilistic perspective, this effect can be interpreted as favoring  semantically plausible regions of the feasible configuration space during geometric solving.

\vspace{-5pt}
\paragraph{Bayesian Formulation.}
Given geometric constraints $C$ (e.g., adjacency, boundary, or continuity) and the semantic priors $\mathcal{P}=\{P_t\}_{t=1}^{T}$ predicted by LASP, the posterior of the reconstructed surface can be expressed as
\begin{equation}
p(S\,|\,H,R)
\propto 
p(C\,|\,S)\;
p(S\,|\,\mathcal{P})\;
p_{\theta}(\mathcal{P}\,|\,H,R),
\end{equation}
where $p(C\,|\,S)$ measures geometric fidelity, $p(S\,|\,\mathcal{P})$ encodes the likelihood of the surface configuration under the semantic priors, and $p_{\theta}(\mathcal{P}\,|\,H,R)$ is the language-conditioned prior distribution learned in Sec.~\ref{subsec:prior}. Maximizing this posterior is equivalent (up to an additive constant independent of $S$) to
\begin{equation}
\label{eq:bayes_energy}
S^*=\arg\min_S
\Big[
E_{\text{fit}}(S;C)
+\lambda\,E_{\text{sem}}(S;\mathcal{P})
\Big],
\end{equation}
where $E_{\text{fit}}=-\log p(C\,|\,S)$ and $E_{\text{sem}}=-\log p(S\,|\,\mathcal{P})$. Rather than defining the exact objective optimized in our implementation, this expression provides an intuitive interpretation of how language-derived priors may bias the feasible configuration space considered during fitting.

\vspace{-5pt}
\paragraph{Theoretical Implication.}
Equation~\ref{eq:bayes_energy} provides a conceptual probabilistic view of how semantic priors may influence geometric solving: rather than changing the solver objective itself, they bias the search toward semantically preferred configurations. In practice, our current framework does not explicitly optimize $E_{\text{sem}}$ inside the geometric kernel. Instead, the engine-level study in Sec.~\ref{subsec:engine_eval} realizes this influence at the configuration level by instantiating discrete prior fields, such as degree, closure, and trimming patterns, as solver-accessible choices.

\begin{figure*}[t]
\centering
\includegraphics[width=\textwidth]{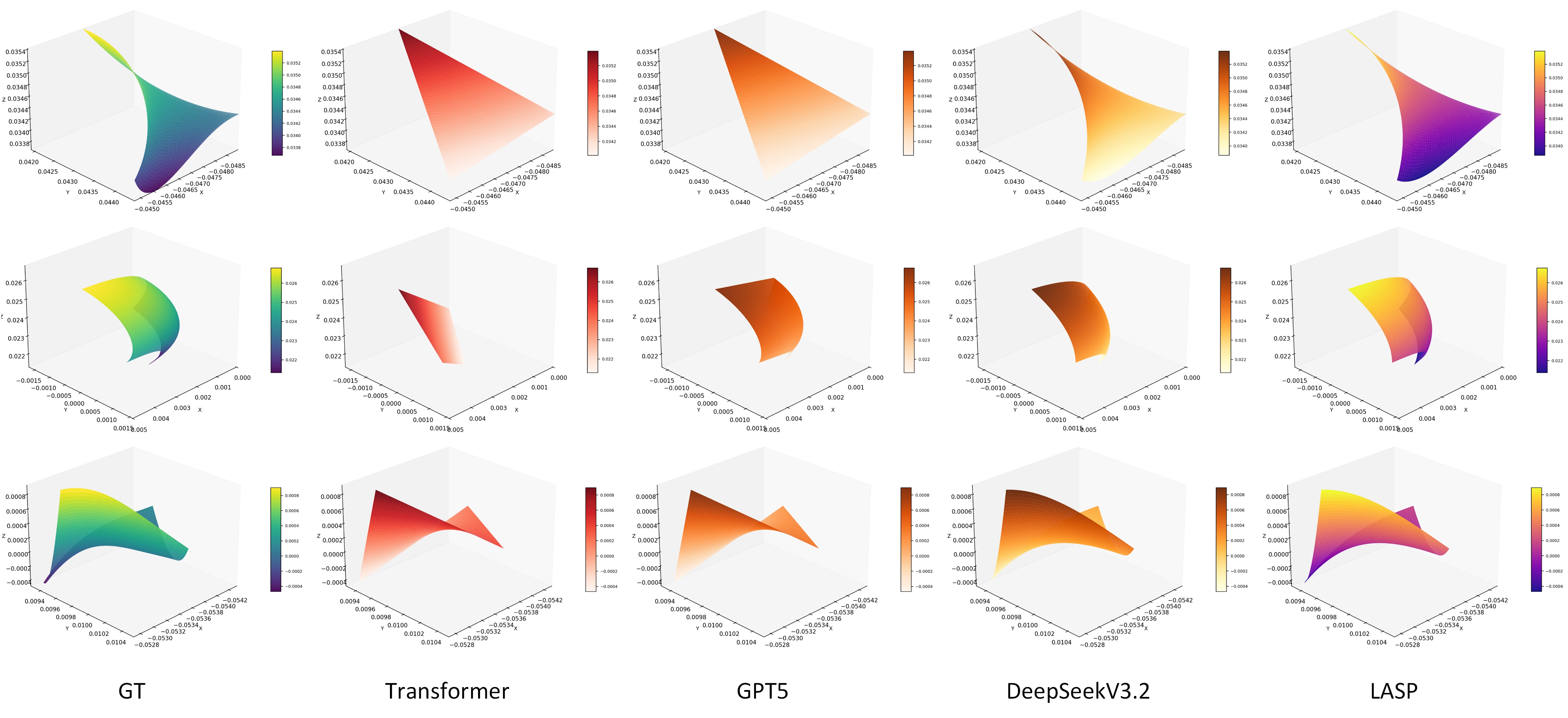}
\caption{\textbf{Qualitative comparison of predicted B-spline parameters.}
All methods reconstruct the surface using their predicted B-spline degrees and pole counts while reusing the ground-truth control points. This isolates the quality of the predicted priors, with LASP producing parameters that generate surfaces most consistent with the reference geometry.}
\label{fig:param_vis}
\vspace{-12pt}
\end{figure*}

\section{Experiments}
\label{sec:experiments}

\subsection{Experimental Setup}
\label{subsec:setup}
\paragraph{Dataset.}
We adopt an extended subset of the ABC dataset~\cite{koch2019abc}, 
which provides paired procedural modeling histories and their corresponding BRep geometries.  
All operations capable of producing B-spline surfaces—primarily \textit{Fillet}, \textit{Chamfer}, and \textit{Shell}—are extracted for Stage~I local learning, so the current evaluation scope is mainly determined by the operation coverage of the dataset.  
After filtering, geometric verification, and controlled data augmentation, 
both Stage~I and Stage~II datasets contain approximately 35,000 valid samples.  
Minor random perturbations of degrees, pole counts, and curvature scales are used to increase sample diversity 
while keeping the statistical distribution consistent with real procedural data.  
Ground-truth B-spline parameters are directly derived from OpenCascade~7.8 to ensure geometric correctness.

\vspace{-5pt}
\paragraph{Implementation.}
LASP is implemented using the \textbf{Qwen3-14B} backbone with full-parameter fine-tuning to enable complete adaptation to geometric reasoning.  
Training is conducted for 4~epochs using the AdamW optimizer (learning rate $2{\times}10^{-5}$, weight decay~0.01) 
on eight NVIDIA~A100 GPUs.  
Each training sample consists of a procedural code sequence and its automatically generated rich-text semantics, 
which are tokenized and concatenated as model input.  
All B-spline reconstructions and quantitative evaluations are executed with OpenCascade, 
including surface trimming, curvature sampling, and distance-based metric computation.

\subsection{Parameter-Level Evaluation}
\label{subsec:param_eval}

Before integrating LASP with any geometric solver, we first assess whether the
model can directly infer geometrically meaningful and semantically coherent
B-spline parameters purely from procedural histories and textual cues.  
This experiment isolates the intrinsic geometric reasoning ability of the language
model—evaluating its capacity to infer surface structure without relying on
any downstream numerical optimization.

We evaluate three complementary aspects: \textit{geometric parameters}
(degrees, pole counts) that reflect surface complexity; \textit{topological flags}
(closure, trimming, rationality) that capture structural regularities; and
\textit{semantic attributes} (continuity class, curvature scale, surface role) 
that indicate whether textual semantics encode design intent. 
Regression metrics use mean absolute error (MAE), while classification is 
evaluated with accuracy and macro-averaged F1.

\vspace{-8pt}
\paragraph{Quantitative Results.}
LASP consistently outperforms all baselines, as shown in
Table~\ref{tab:param_results}. It achieves $F_1{=}0.95$ in B-spline detection
and a mean degree accuracy of 0.85, indicating strong discriminability across
geometric scales and structures. On higher-level semantics, LASP surpasses GPT-5\cite{achiam2023gpt}
and DeepSeek-v3.2\cite{liu2024deepseek} on both \textit{ContinuityClass} ($F_1{=}0.698$) and
\textit{SurfaceRole} ($F_1{=}0.979$), confirming that linguistic signals provide
effective priors for inferring design functionality. Although some rare
topology categories (e.g., \textit{UClosed}, $<5\%$ frequency) yield lower
F1-scores, their high accuracies indicate that LASP avoids false positives even
under extreme imbalance. For brevity, we use the following abbreviations in
Table~\ref{tab:param_results}: 
\textbf{BSpl-Det} (B-spline detection F1), 
\textbf{Deg-Acc} (degree accuracy U/V/mean), 
\textbf{Pole-MAE} (pole-count MAE U/V/total), 
\textbf{Rat-F1} (rationality flags U/V), 
\textbf{Topo} (topology flag Acc/F1), 
\textbf{Cont} (continuity class MF1/Acc), 
\textbf{Curv} (curvature scale MF1/Acc), 
and \textbf{Role} (surface role MF1/Acc).
Controlled attribution under matched settings, including a matched history-only variant, is further analyzed in Sec.~\ref{subsec:ablation}.

\vspace{-8pt}
\paragraph{Qualitative Visualization.}
To qualitatively assess the plausibility of predicted parameters, 
we reconstruct surfaces using the degrees and pole configurations generated by each model 
while keeping ground-truth control points fixed to eliminate solver effects.  
Figure~\ref{fig:param_vis} compares representative samples across 
the reference ground truth (GT), Transformer baseline, GPT-5, DeepSeek-v3.2, and LASP.  
Surfaces predicted by LASP exhibit the highest geometric fidelity, 
capturing fine curvature transitions and local continuity consistent with the ground truth.  
This demonstrates that LASP’s predicted priors encode semantically consistent and geometrically plausible structures, 
providing a solid foundation for the solver-level analysis in Sec.~\ref{subsec:engine_eval}.

\begin{table}[t]
\centering
\caption{\textbf{Engine-Level Evaluation via Solver Configuration Instantiation.}
B-spline fitting reconstruction error under different solver-accessible parameterization settings.}
\label{tab:engine_eval}
\begin{tabular}{lccc}
\toprule
\textbf{Config} & \textbf{RMS↓} & \textbf{HD↓} & \textbf{Med↓} \\
\midrule
Default & 0.0491 & 0.2746 & 0.0193 \\
\textbf{+ Semantic Priors} & \textbf{0.0141} & \textbf{0.0395} & \textbf{0.0103} \\
\bottomrule
\end{tabular}
\vspace{-8pt}
\end{table}
\subsection{Engine-Level Evaluation via Solver Configuration Instantiation.}
\label{subsec:engine_eval}

Modern geometric kernels encapsulate operation-level parameterization,
making it difficult to directly inject semantic priors, such as control-point
initialization or knot placement.
To evaluate whether LASP’s predicted semantics can nevertheless provide practical guidance for downstream fitting, we instantiate discrete semantic priors into solver-accessible configuration choices for the B-spline fitting process, rather than directly modifying the kernel optimization loop.
Surfaces generated by extrusion, revolution, or blending operations are fitted
using shape-consistent parameter settings, while the default configuration
relies on the kernel’s built-in parameters.
Although the numerical kernel itself remains unchanged, the resulting fitting setup is biased toward semantically consistent parameter configurations, providing a practical instantiation of intent-aware configuration for downstream surface fitting.

We employ three complementary metrics to assess geometric fidelity:
RMS distance\cite{cignoni1998metro} captures overall fitting quality,
Hausdorff distance\cite{aspert2002mesh} highlights the worst-case deviation, and
Median distance\cite{seitz2006comparison} provides robustness against outliers.
Together, these metrics probe global accuracy, local stability, and robustness of the fitted surfaces.

As summarized in Table~\ref{tab:engine_eval}, the semantically guided fitting achieves 
significantly lower reconstruction errors across all metrics, 
reducing RMS deviation by 71.2\% and Hausdorff distance by 85.6\% ($p{<}0.001$). 
These results demonstrate that semantic guidance can effectively improve downstream fitting quality through solver-accessible configuration choices that are more consistent with the procedural-semantic context, serving as empirical evidence that such priors can act as useful inductive biases without modifying solver internals.

\begin{figure}[t]
\centering
\includegraphics[width=\linewidth]{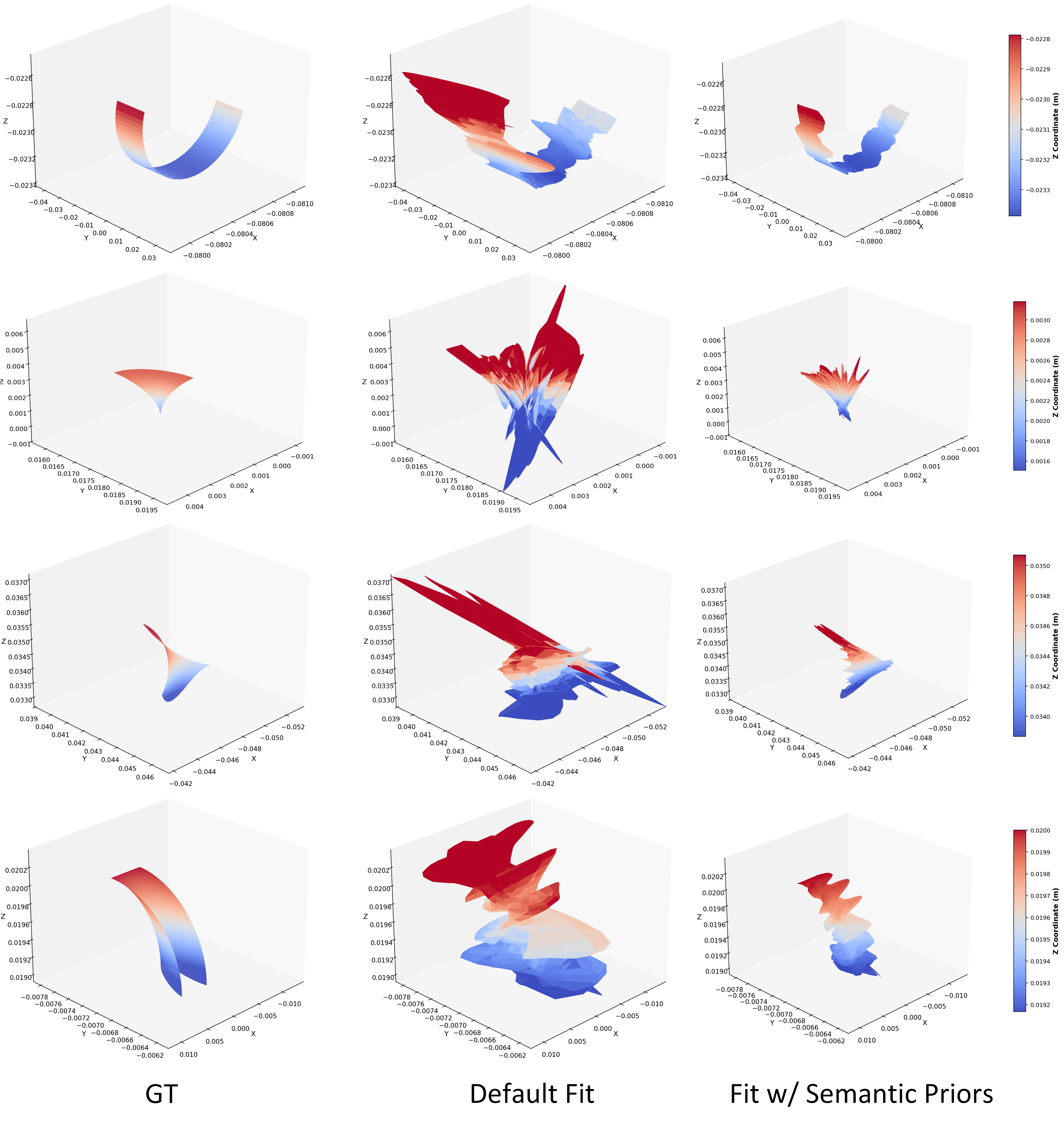}
\caption{\textbf{Qualitative comparison of fitted B-spline surfaces.}
Representative examples from extrusion, revolution, and blending operations.
Columns show the ground truth surface, the default fitting configuration, and fitting guided by LLM-predicted semantic priors.
Color shading indicates the $z$-axis height of the reconstructed surface.}
\label{fig:engine_vis}
\vspace{-5mm}
\end{figure}

Figure~\ref{fig:engine_vis} provides qualitative comparisons across different
operation types.
Each triplet shows the reference ground truth, the default fitting result, and
the fitting with LASP-derived semantic priors. 
Unlike Fig.~\ref{fig:param_vis}, which visualizes parameter-level predictions under fixed control points, 
this figure directly shows solver-generated surfaces where the color variation reflects the surface height along the $z$-axis.  
Semantic priors yield smoother and more coherent reconstructions, particularly in high-curvature regions, confirming that operation-level semantics provide meaningful configuration guidance for the fitting process.

\subsection{Ablation Study}
\label{subsec:ablation}

To examine the contribution of each component within LASP and to better attribute the source of its gains, 
we conduct controlled ablation experiments together with a matched history-only comparison. 
Specifically, we evaluate a history-only variant under the same Qwen3-14B fine-tuning protocol, 
as well as variants that remove Stage~I geometric pretraining, data augmentation, or rich-text semantics, respectively.  
All models are trained under identical optimization configurations and evaluated using the
same parameter-level metrics as in Sec.~\ref{subsec:param_eval}.

As shown in Table~\ref{tab:ablation}, each component plays a distinct role. 
The matched Qwen3-14B history-only variant already falls clearly behind the full LASP model
(Pole-MAE: 0.61 vs.\ 0.56; \textit{ContinuityClass} $F_1$: 0.59 vs.\ 0.70; 
\textit{SurfaceRole} $F_1$: 0.86 vs.\ 0.98), indicating that the gains cannot be explained by backbone scale or fine-tuning alone and that semantic conditioning is essential. 
Removing rich-text semantics within the two-stage LASP design yields the largest degradation in high-level
reasoning (\textit{SurfaceRole} $F_1$: 0.98 $\rightarrow$ 0.91; 
\textit{ContinuityClass} $F_1$: 0.70 $\rightarrow$ 0.62), confirming that operation-level
semantic context is crucial for capturing high-level procedural semantics. 
Disabling data augmentation produces similar declines, suggesting that moderate procedural diversity is
necessary for robust generalization. Removing Stage~I geometric pretraining
primarily affects numerical precision (Pole-MAE: 0.56 $\rightarrow$ 0.66), confirming that
explicit geometric grounding stabilizes quantitative inference. 
Overall, LASP’s performance emerges from the combined effects of geometric
supervision, semantic conditioning, and data diversity.

\begin{table}[t]
\centering
\setlength{\tabcolsep}{0.5pt}
\caption{\textbf{Ablation and controlled attribution analysis.} 
 Key metrics showing the effect of a matched history-only variant under the same Qwen3-14B fine-tuning protocol, together with the effects of rich-text semantics, data augmentation, and Stage~I pretraining.}
 \vspace{-5pt}
\label{tab:ablation}
\begin{tabular}{lccc}
\toprule
\textbf{Variant} & \textbf{Pole MAE↓} & \textbf{Cont F1↑} & \textbf{Role F1↑} \\
\midrule
\textbf{Full LASP} & \textbf{0.56} & \textbf{0.70} & \textbf{0.98} \\
Qwen3-14B(history-only) & 0.61 & 0.59 & 0.86 \\
w/o Rich-Text Sem. & 0.59 & 0.62 & 0.91 \\
w/o Augment. & 0.58 & 0.63 & 0.92 \\
w/o Stage~I Pretrain. & 0.66 & 0.68 & 0.95 \\
\bottomrule
\end{tabular}
\vspace{-5mm}
\end{table}

\section{Conclusion}
\label{sec:conclusion}

We introduced \textbf{LASP}, a language-augmented framework that brings contextual semantic reasoning into B-spline surface construction. Rather than modifying geometric kernels, LASP operates as an external semantic layer that transforms procedural modeling histories into rich-text descriptions and uses these to predict structured B-spline priors. These priors provide high-level inductive signals that guide conventional B-spline fitting toward more coherent and design-aligned solutions. A two-stage learning strategy allows LASP to capture both local geometric regularities and global contextual dependencies, producing priors that consistently improve surface quality and parameter stability. The results demonstrate that language-driven reasoning provides a powerful inductive bias for geometric solving, pointing toward deeper connections between symbolic intent and numerical optimization.

{
    \small
    \bibliographystyle{ieeenat_fullname}
    \bibliography{main}
}


\end{document}